\documentclass[10pt,twocolumn,letterpaper]{article}

\usepackage{cvpr}              

\definecolor{cvprblue}{rgb}{0.21,0.49,0.74}
\usepackage{enumitem}
\usepackage[pagebackref,breaklinks,colorlinks,allcolors=cvprblue]{hyperref}
\usepackage[accsupp]{axessibility} 
\def\paperID{16549} 
\def\confName{CVPR}
\def\confYear{2025}

\title{UniCompress: Token Compression for Unified Vision–Language \\ Understanding and Generation}

\author{
Ziyao Wang$^{1,2}$ \ Chen Chen$^{1}$ \ Jingtao Li$^{1}$ \ Weiming Zhuang$^{1}$ \ Jiabo Huang$^{1}$  \\ Ang Li$^{2}$ \ Lingjuan Lyu$^{1,}$\thanks{Correspondence to: Lingjuan Lyu \textless Lingjuan.lv@sony.com\textgreater} \\
{\small$^{1}$Sony AI \quad
$^{2}$University of Maryland, College Park}
}

\newcommand{\n}{\textsc{UniCompress}}

\begin{document}
\maketitle
\begin{abstract}
Unified models aim to support both understanding and generation by encoding images into discrete tokens and processing them alongside text within a single autoregressive framework. This unified design offers architectural simplicity and cross-modal synergy, which facilitates shared parameterization, consistent training objectives, and seamless transfer between modalities. However, the large number of visual tokens required by such models introduces substantial computation and memory overhead, and this inefficiency directly hinders deployment in resource constrained scenarios such as embodied AI systems. In this work, we propose a unified token compression algorithm {\n} that significantly reduces visual token count while preserving performance on both image understanding and generation tasks. Our method introduces a plug-in compression and decompression mechanism guided with learnable global meta tokens. The framework is lightweight and modular, enabling efficient integration into existing models without full retraining. Experimental results show that our approach reduces image tokens by up to \(4\times\), achieves substantial gains in inference latency and training cost, and incurs only minimal performance degradation, which demonstrates the promise of token-efficient unified modeling for real world multimodal applications.

\end{abstract}    
\section{Introduction}
\label{sec:intro}


Recent research on multimodal learning has been moving towards unified models that can \emph{understand} and \emph{generate} images within a single autoregressive framework~\cite{wu2024vila,ma2025unitok,zhuang2025vargpt}. A common solution is to encode images into discrete visual tokens produced by a learned tokenizer, then feeds these tokens, together with text tokens, into a language-model backbone. This shared token space enables a broad spectrum of multimodal tasks (\eg, image captioning~\cite{byun2025unifying}, VQA~\cite{zhang2025unified}, image editing~\cite{bai2025uniedit}) to be handled by one architecture, simplifying deployment and multi-task training.

\begin{figure}[t]
  \centering
  \includegraphics[width=\linewidth]{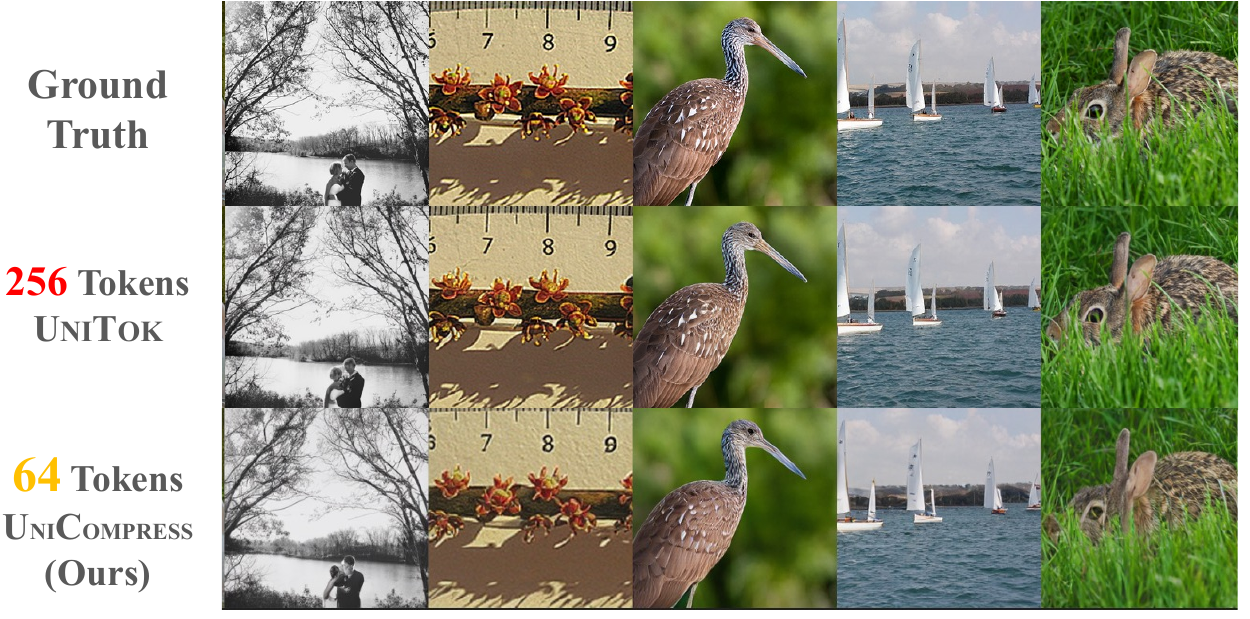}
  \vspace{-0.5cm}
  \caption{We propose {\n}, a plug-in-and-play token compression algorithm for unified models. The samples are from UniTok~\cite{ma2025unitok}.}
  \label{fig:compression_ratio}
\end{figure}

However, a practical limitation of these unified models is \emph{token efficiency}. Tokenizers from the discrete codebook family (\eg, VQ\mbox{-}VAE~\cite{razavi2019generating}, the dVAE used in DALL\(\cdot\)E~\cite{ramesh2021zero}, and VQGAN~\cite{esser2021taming}) often map a \(512\times512\) image to \(32\times32=1024\) tokens (\ie downsampling by a factor of 16 along height and width).
Unified models must then use long visual sequences for understanding and adopt equally long sequences for generation, increasing memory footprint, training cost, and inference latency. Sharing a single visual tokenizer across understanding and generation~\cite{ma2025unitok,jiao2025unitoken} reduces engineering complexity but does not reduce the sequence-length. A straightforward solution is to compress the visual tokens. However, experimental results demonstrate that \textbf{naïve downsampling or uniform token pruning, while effective for image understanding, significantly degrades generation performance by more than 15\%}. This sets the main challenge since image generation relies on fine-grained, spatially consistent tokens to accurately reconstruct details, hence is sensitive to pruning.



Another challenge is ensuring low-cost. While we can train another tokenizer with better token efficiency (e.g. one-D tokenizer) to replace the existing one, but doing so often requires downstream language model training from scratch, which is costly. A more practical approach should be modular, allowing seamless integration with existing tokenizers without the need for full retraining. Thus, the secondary challenge is \emph{how to develop a plugin-based compression method compatible with any tokenizer, avoiding expensive LLM retraining.}

To address the challenges, we introduce {\n}, a plug-in token compression framework that reduces visual tokens by up to \(4\times\) while maintaining quality for both understanding and generation. Motivated by~\cite{jiang2024lemevit,harvill2025lossless}, {\n} inserts two lightweight modules around an off-the-shelf discrete tokenizer: (i) a \emph{compressor} that converts a dense \(H\times W\) token grid into a short sequence of compressed tokens \emph{augmented by a small set of learnable global meta tokens} capturing holistic semantics; and (ii) a \emph{global-guided decompressor}
that reconstructs high-fidelity token grids conditioned on those global tokens, restoring long-range structure. 
We first train the tokenizer together with the compressor and decompressor for image reconstruction task, then freeze the compressed tokenizer and lightly finetune the language model on the compressed tokens for both understanding and generation. 
The result is a more efficient unifed model with more compact input/output sequence, while retaining the understanding and generation performance through global-guided decompression.
\footnote{We instantiate a fixed compression ratio for controlled comparisons; the design naturally extends to content-adaptive rates (see \S\ref{sec:method}).}
Intuitively, the compressed tokens carry salient local evidence, whereas the global meta tokens provide scene-level constraints. During generation, the decompressor uses these global tokens as semantic anchors to autoregressively refine local textures and boundaries, mitigating the detail loss observed with uniform token compression. 

On standard understanding and generation benchmarks, {\n} reduces visual tokens by \(4\times\) (for example, \(256\rightarrow64\)) 
while keeping performance within small margins ($\leq$ 3-pt drop on understanding; $\leq$ 5-pt FID increase on generation.
Relative to uncompressed baselines, {\n} yields up to 41.8\% lower inference latency and 15.4\% shorter training time, together with substantial latency savings from shorter sequences. These results indicate a practical path to unified models suitable for resource-constrained platforms. 
Our key contributions are as follows:

\begin{itemize}
  \item We highlight token efficiency as a bottleneck in unified models and show that naïve token compression disproportionately harms \emph{generation}. We formalize the objective of a single compact visual token space usable for both understanding and generation.
  \item We propose \n, a plug-in compression framework with \emph{global-guided} autoregressive decompression. It shortens the visual sequence while preserving generation detail and seamlessly integrates into existing unified models.
  \item We demonstrate strong empirical results across different unified models. Our method achieves up to a 4$\times$ token reduction while maintaining competitive performance. on both understanding and generation tasks. For both understanding and generation tasks, we cap the performance drop at $\leq$ 5\%, and on some benchmarks we fully match pre-compression performance.
\end{itemize}

\section{Related Work}
\label{sec:related}

\paragraph{Unified Foundation Model.}
Recent advances in multi-model learning have led to vision foundation models \cite{zhuang2025argus} and unified multimodal models \cite{wu2024vila,ma2025unitok} that support both visual understanding and image generation within a single framework. These models typically encode images into discrete token sequences and process them alongside text using a language model backbone. \textsc{DreamLLM}~\cite{dong2023dreamllm} treats image and text as a joint sequence and leverages autoregressive modeling to seamlessly compose multimodal content. \textsc{VILA-U}~\cite{wu2024vila} moves away from diffusion-based generation, opting instead for a fully token-level decoder that unifies captioning and image synthesis via next-token prediction. To address the limitations of low-capacity tokenizers, \textsc{UniTok}~\cite{ma2025unitok} expands the expressiveness of visual tokens using a multi-codebook quantizer, balancing semantic abstraction and reconstruction detail. \textsc{VARGPT}~\cite{zhuang2025vargpt} takes a hierarchical approach by predicting both content and resolution scale, enabling controllable and efficient image generation over multiple granularities. Meanwhile, \textsc{UniFork}~\cite{li2025unifork} questions the viability of fully shared backbones and proposes a Y-shaped design that splits deeper layers by task to reduce interference while preserving early fusion. Other works utilize diffusion model as the generation head. For instance, \textbf{OpenUni}~\cite{wu2025openuni} and \textbf{Bagel}~\cite{deng2025emerging} further push token-level autoregressive generation with diffusion models. To facilitate joint training and architectural simplicity, many of these models adopt a shared tokenizer across modalities, often using ViT-style patch embeddings or VQ-based discrete representations (e.g., dVAE or VQ-GAN)~\cite{xie2025muse,zhang2025unified}.

However, representing each image with hundreds or even thousands of tokens introduces substantial computational overhead. This limitation highlights the need for a unified compression framework that can reduce token redundancy while preserving task performance.

\paragraph{Visual Token Compression of Foundation Model.}
A key bottleneck in implementing VLMs lies in the large number of discrete tokens used to represent images. Each image is typically encoded into hundreds or even thousands of tokens, which significantly increases sequence length and leads to high memory usage, latency, and computational costs—particularly during inference. Current token compression works mainly focus on image understanding tasks. \cite{alvar2025divprune,sun2025lvpruning,ye2025voco,wang2025dynamic,li2024inference} prune image tokens either before feeding them into the LLM or within the LLM layers based on attention scores. These understanding-oriented pruning methods aim to reduce pipeline FLOPs without fine-tuning, thereby improving inference efficiency. For image generation models \cite{sehwag2025stretching}, MaskGIT~\cite{chang2022maskgit} reduces inference latency through masked token prediction. \cite{tian2024visual} improves image generation efficiency by reformulating autoregressive decoding as next-scale prediction instead of next-token prediction. HMAR~\cite{kumbong2025hmar} further combines next-scale prediction with masked generation to enhance decoding speed. However, inference-time pruning or prediction optimization techniques are typically task-specific and cannot be directly applied to both understanding and generation tasks in a unified setting~\cite{zhang2024sparsevlm}. Other works such as TiTok~\cite{yu2024image} uses a 1D tokenizer to compress visual tokens. While this works well for understanding tasks, it underperforms on generation tasks due to the loss of spatial information. This highlights the need for an efficient unified token compression framework. 
\section{Our Method: {\n}}
\label{sec:method}

\begin{figure*}[htbp]
    \centering
    \includegraphics[width=0.95\textwidth]{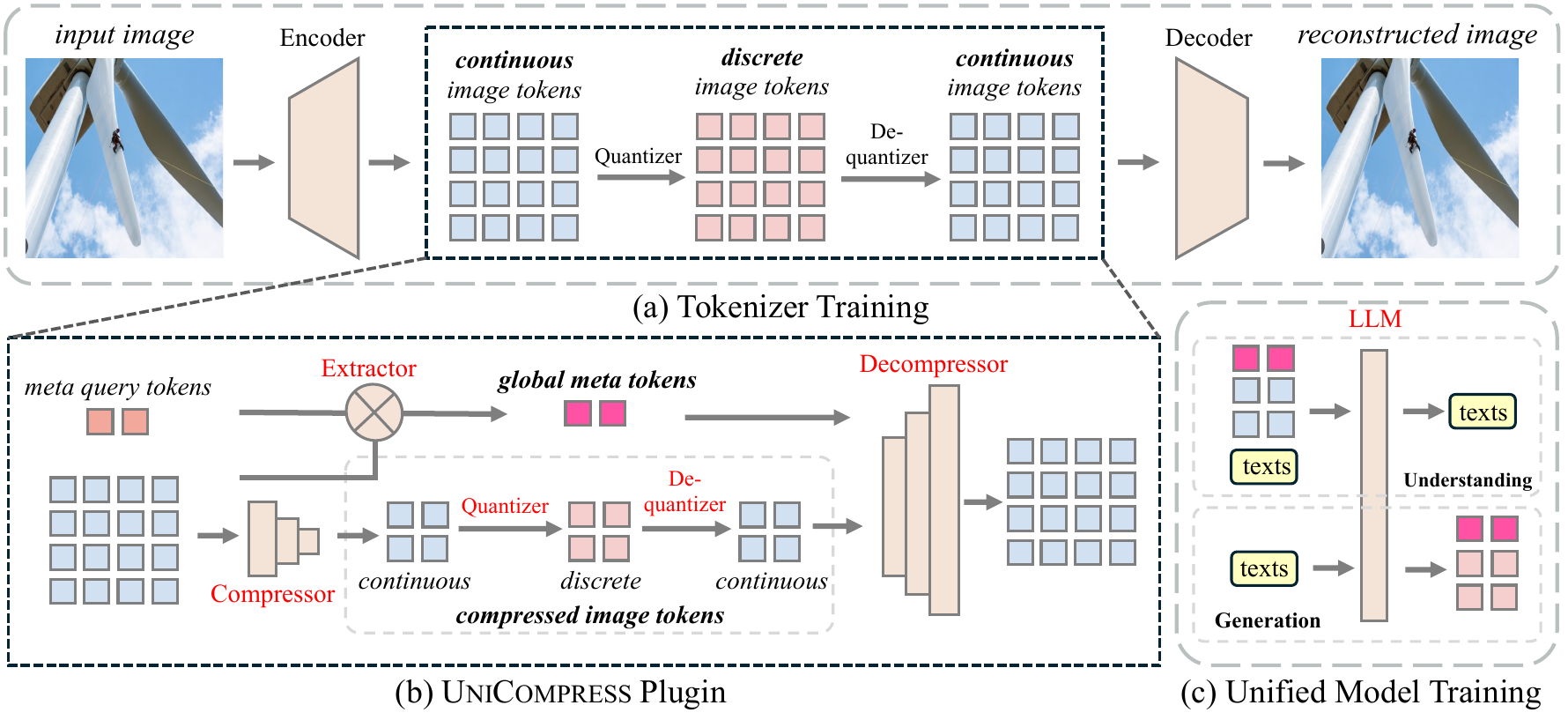}
    \caption{Overview of {\n}. The tokenizer is augmented with three modules: a global token extractor, a token compressor, and an autoregressive decompressor. The language model consumes a compact visual sequence for understanding and produces compressed-domain targets for generation.}
    \label{fig:overview}
\end{figure*}

\subsection{Overview}
As shown in Fig.~\ref{fig:overview}, {\n} augments the visual tokenizer with three lightweight modules while keeping the LLM unchanged: a global-token extractor that uses one-way cross-attention to summarize scene-level semantics; a pooling-based compressor that reshapes the token grid and aggregates non-overlapping patches (e.g., \(2{\times}2\), \(4{\times}4\)) into a shorter sequence
and an autoregressive decompressor that later expands the compact representation back to a dense token grid for the image decoder.
During inference, for image understanding tasks, we place the visual subsequence into the LLM input (after a linear projection to the LLM embedding space). For image generation, the LLM predicts global tokens and compressed visual tokens, we then feed them to the codebook and decompressor to recover the final image. 
Training is done in two stages: we first train the tokenizer with {\n} modules with a reconstruction loss; then we freeze the compressed tokenizer and mildly finetune the LLM model. This “compress once, reuse for both”
interface reduces sequence length and compute while preserving global structure. Moreover, our approach readily adapts to other unified model designs, including configurations with multiple image tokenizers or diffusion models.

\subsection{Global Token Extraction via Cross-Attention}

Let the base encoder output a continuous sequence of visual tokens $ \mathbf{X}\in\mathbb{R}^{T\times d} $ where $ T=H\times W $ and $ d $ is the embedding size. We introduce a small set of learnable meta query tokens $ \mathbf{Q}\in\mathbb{R}^{N_g\times d} $ that extract global context from $ \mathbf{X} $ using one-way cross-attention: meta tokens query the full image-token field, while image tokens pass through unchanged. For each image, the image-specific global tokens $ \mathbf{G}\in\mathbb{R}^{N_g\times d} $ are computed as
\begin{equation}
\mathbf{G}=\mathrm{MHA}\big(\mathbf{Q}W_Q,\ \mathbf{X}W_K,\ \mathbf{X}W_V\big),
\end{equation}
where $ W_Q,W_K,W_V\in\mathbb{R}^{d\times d} $ are learned projections and $ \mathrm{MHA} $ denotes multi-head attention. We apply residual and normalization on the meta branch,
\begin{equation}
\mathbf{G}\leftarrow \mathrm{LN}\!\big(\mathbf{Q}+\mathbf{G}\big).
\end{equation}
In practice, $ N_g $ is much smaller than $ T $, so the additional sequence length is minor while providing strong global guidance for layout and object relations. Global tokens use their own learned positional embeddings.

\subsection{Image Token Compression via Average Pooling}

To shorten the visual sequence, we aggregate local tokens within non-overlapping spatial patches. We first reshape $\mathbf{X}$ back to its $H{\times}W$ grid using the tokenizer's canonical rasterization. 
The default operation is fixed-size average pooling applied on the embedding field, which reduces spatial redundancy while preserving coarse structure. Given a downsampling factor $ s $, the compressed continuous sequence is
\begin{equation}
{\hat{\mathbf{X}}}^{\text{cont}}=\mathrm{AvgPool}(\mathbf{X}, s),\qquad \tilde T=T/s^{2}.
\end{equation}
We then insert the compressed visual segment into the multimodal sequence using three learned special embeddings: an image-begin token $[\texttt{IMG\_BOS}]$, a separator token $[\texttt{IMG\_SEP}]$ that splits global and local tokens, and an image-end token $[\texttt{IMG\_EOS}]$. For understanding tasks (e.g., image captioning, VQA), depending on the unified model’s design, the language model consumes either continuous tokens or discrete tokens. 

For the generation task, the tokenizer includes a discrete codebook of size $ K $. We use the original quantizer to quantize compressed global and local tokens for image generation training.
The indices have spatial size $ H/s\times W/s $ and take values in $ \{1,\dots,K\} $. We use the same codebook for global and local streams:
\begin{equation}
\hat{\mathbf{Z}}^{(g)} \in \{1,\dots,K_g\}^{N_g},\qquad
\hat{\mathbf{Z}}^{(x)} \in \{1,\dots,K_x\}^{\tilde T}.
\end{equation}
This dual representation allows the same compression mechanism to support continuous features for understanding and discrete targets for generation. Image and text tokens share positional embeddings. The target sequence is ordered as
\begin{equation}
\texttt{[IMG\_BOS]},\;\hat{\mathbf{Z}}^{(g)}_{1{:}N_g},\;
\texttt{[IMG\_SEP]},\;\hat{\mathbf{Z}}^{(x)}_{1{:}\tilde T},\;
\texttt{[IMG\_EOS]}.
\end{equation}

\subsection{Autoregressive Decompression Guided by Global Tokens}

In generation, the language model autoregressively outputs both the global meta tokens and the compressed local tokens in the discrete domain. These indices are mapped back to continuous compressed features by codebook lookups,
\begin{equation}
\hat{\mathbf{G}} = \mathcal{E}\!\big(\hat{\mathbf{Z}}^{(g)}\big),
\qquad
{\hat{\mathbf{X}}}^{\text{deq}} = \mathcal{E}\!\big(\hat{\mathbf{Z}}^{(x)}\big),
\end{equation}
where $\mathcal{E}$ denotes the codebook. 

Given $(\hat{\mathbf{G}}, {\hat{\mathbf{X}}}^{\text{deq}})$, the decompressor expands the compact representation into a dense sequence of continuous tokens at the original resolution expected by the image decoder. We implement $f_{\mathrm{dec}}$ as a Transformer decoder with masked self-attention over the generated dense prefix and cross-attention to $(\hat{\mathbf{G}}, {\hat{\mathbf{X}}}^{\text{deq}})$ at every layer. At raster step $t$, the next dense token $\mathbf{x}_t$ is predicted by
\begin{equation}
\mathbf{x}_t = f_{\mathrm{dec}}\!\big(\mathbf{X}^{\text{dense}}_{<t},\ {\hat{\mathbf{X}}}^{\text{deq}},\ \hat{\mathbf{G}}\big),
\end{equation}
where $\mathbf{X}^{\text{dense}}_{<t}$ are previously generated dense tokens and causal masking is applied along the generation order.

Training uses teacher forcing against the tokenizer’s dense targets $\mathbf{X}$. The reconstruction objective combines a token-level regression term in the dense feature space with a codebook consistency term:
\begin{equation}
\mathcal{L}_{\mathrm{recon}} = \mathcal{L}_{\mathrm{reg}} + \lambda_{\mathrm{cb}}\mathcal{L}_{\mathrm{cb}}.
\end{equation}

\subsection{Lightweight Training Pipeline}

The overall procedure is modular and keeps changes on the tokenizer side.

\textbf{Stage one (tokenizer-side training).} We freeze the LLM and train the tokenizer stack with the reconstruction objective $\mathcal{L}_{\mathrm{recon}}$. The stack includes: the learnable meta-query global extractor (producing $\mathbf{G}$), the fixed average-pooling compressor (producing ${\hat{\mathbf{X}}}^{\text{cont}}$), the codebook $\mathcal{E}$, and the decompressor $f_{\mathrm{dec}}$. This stage learns to map a dense sequence $\mathbf{X}$ to a compact pair $(\mathbf{G},{\hat{\mathbf{X}}}^{\text{cont}})$ and back to dense tokens with high fidelity.

\textbf{Stage two (LLM training).} We freeze the tokenizer and train the LLM on compact data. For understanding, the LLM consumes continuous tokens $\{\mathbf{G},{\hat{\mathbf{X}}}^{\text{cont}}\}$. For generation, the LLM autoregressively outputs the discrete indices for both streams; the discrete tokens are then de-quantized via $\mathcal{E}_g,\mathcal{E}_x$ to $(\hat{\mathbf{G}},{\hat{\mathbf{X}}}^{\text{deq}})$ and expanded by $f_{\mathrm{dec}}$ to dense tokens for the image decoder. Because the LLM interface is a standard autoregressive sequence over special tokens and visual indices, {\n} integrates into existing unified model backbones without architectural modification.


\begin{table*}[htp] 
\centering 
\caption{Unified model performance on visual understanding benchmarks (higher is better).
\textsc{XXX-Compressed} denotes the same backbone with our plug-in token compression ($s{=}2$, $N_g{=}4$). 'MME Cog.' refers to the score of MME Cognition.}
\vspace{-0.3cm}
\resizebox{0.9\textwidth}{!}{ 
\begin{tabular}{l|cccccccc} 
\toprule 
\textbf{Methods} & \textbf{GQA} & \textbf{MME Cog.} & \textbf{MME} & \textbf{POPE} & \textbf{Seed-bench} & \textbf{TextVQA} & \textbf{MMMU} & \textbf{MM-Bench} \\
\midrule 
\textsc{UniTok}  & 55.71 & 251.79 & 1162.50 & 82.66 & 49.22 & 24.90 & 26.22 & 40.34 \\ 
\textsc{UniTok-Compressed}  & 53.07 & 235.00 & 1036.05 & 79.36 & 48.07 & 24.66 & 26.22 & 42.14 \\
\midrule 
\textsc{Vila-U} & 53.43 & 247.65 & 1130.20 & 81.45 & 47.88 & 24.11 & 25.72 & 39.55 \\ 
\textsc{Vila-U-Compressed}  & 51.90 & 241.24 & 1105.35 & 80.75 & 47.30 & 23.85 & 25.55 & 39.94 \\ 
\midrule 
\textsc{VARGPT} & 58.12 & 269.30 & 1290.65 & 88.04 & 50.54 & 26.18 & 27.13 & 44.22 \\ 
\textsc{VARGPT-Compressed}  & 55.90 & 265.83 & 1272.80 & 84.99 & 48.41 & 25.34 & 27.40 & 41.15 \\ 
\midrule 
\midrule 
\textsc{UniFork} & 54.20 & 246.51 & 1120.25 & 81.22 & 48.82 & 24.36 & 26.22 & 39.84 \\ 
\textsc{UniFork-Compressed} & 50.10 & 232.86 & 1045.05 & 78.97 & 47.95 & 24.89 & 25.90 & 40.50 \\
\midrule 
\textsc{OpenUni} & 53.33 & 243.59 & 1108.70 & 80.60 & 48.39 & 24.30 & 25.80 & 39.28 \\ 
\textsc{OpenUni-Compressed} & 49.80 & 230.21 & 1030.10 & 78.17 & 47.51 & 23.10 & 25.72 & 39.88 \\
\midrule 
\textsc{BAGEL} & 60.05 & 277.80 & 1312.40 & 89.20 & 51.10 & 26.90 & 34.05 & 45.10 \\
\textsc{BAGEL-Compressed} & 59.10 & 274.50 & 1304.10 & 88.60 & 50.80 & 36.50 & 27.80 & 44.50 \\
\bottomrule 
\end{tabular} 
} 
\label{tab:understanding-performance} 
\end{table*}

\begin{table}[t] 
\centering 
\caption{Performance of the original and compressed unified models on image generation benchmarks. \textsc{XXX-Compressed} inserts {\n} without changing the LM interface. Lower FID and higher CLIP indicate better quality.} 
\vspace{-0.3cm}
\resizebox{0.78\linewidth}{!}{
\begin{tabular}{l|cc} \toprule \textbf{Methods} & \textbf{FID($\downarrow$)} & \textbf{CLIP($\uparrow$)} \\
\midrule 
\textsc{UniTok} & 16.14 & 30.5 \\
\textsc{UniTok-Compressed} & 16.33 & 25.0 \\
\textsc{Vila-U} & 14.80 & 29.8 \\
\textsc{Vila-U-Compressed} & 16.37 & 28.9 \\
\textsc{VARGPT} & 14.77 & 24.2 \\
\textsc{VARGPT-Compressed} & 15.02 & 21.6 \\
\midrule
\textsc{UniFork} & 17.83  &25.5  \\      
\textsc{UniFork-Compressed}  &20.24 &26.0  \\   
\textsc{OpenUni} & 16.45 &26.7 \\         
\textsc{OpenUni-Compressed}   &24.29 & 22.3 \\ 
\textsc{BAGEL} & 12.73 &32.0 \\         
\textsc{BAGEL-Compressed}   &17.22 & 28.8 \\ 
\bottomrule 
\end{tabular}}
\label{tab:generation-performance} 
\end{table}

\begin{table*}[t] 
\centering \caption{Wall-clock time with/without plug-in token compression. 
Understanding: ShareGPT4V\_PT (train), GQA (inference). 
Generation: JDB (train), MJHQ-30K (inference). Lower is better. Although the model is trained on the two datasets jointly in other experiments, the training times in this table were measured by training on each dataset separately.
} 
\vspace{-0.3cm}
\resizebox{0.8\textwidth}{!}{ \begin{tabular}{l|cccc} 
\toprule \textbf{Methods} & \textbf{Und. Train (h)} & \textbf{Gen. Train (h)} & \textbf{Und. Inference (min)} & \textbf{Gen. Inference (min)} \\
\midrule \textsc{UniTok} & 0.94 & 4.60 & 5.41 & 32.25 \\
\textsc{UniTok-Compressed} & \textbf{0.81} & \textbf{3.89} & \textbf{5.25} & \textbf{18.96} \\
\midrule \textsc{Vila-U} & 0.89 & 4.25 & 5.12 & 30.15 \\
\textsc{Vila-U-Compressed} & \textbf{0.77} & \textbf{3.65} & \textbf{5.09} & \textbf{17.55} \\
\midrule \textsc{VARGPT} & 1.13 & 5.52 & 5.45 & 38.74 \\
\textsc{VARGPT-Compressed} & \textbf{0.92} & \textbf{4.67} & \textbf{5.43} & \textbf{22.71} \\
\midrule \textsc{UniFork} & 0.90 & 4.30 & 5.82 & 29.90 \\
\textsc{UniFork-Compressed} & \textbf{0.77} & \textbf{3.70} & \textbf{5.80} & \textbf{17.60} \\
\bottomrule 
\end{tabular}}
\label{tab:efficiency}
\end{table*} 

\section{Experiments} \label{sec:experiments} We evaluate our unified token compression framework on a wide range of multimodal understanding and generation tasks. The experiments are designed to answer the following questions: 

\begin{enumerate}[label=\textbf{Q\arabic*}, leftmargin=2em, labelsep=0.6em]
  \item Can our method preserve vision-language understanding accuracy under visual token compression?
  \item Does the generation performance remain competitive when decoding from compressed image tokens?
  \item How much efficiency gain (training time, inference time, and FLOPs) does compression bring?
\end{enumerate}

\subsection{Experimental Setup} %
\paragraph{Models and Datasets.}
We adopt Llama-3.2-1B~\cite{grattafiori2024llama} as the language backbone.
For pre-training, we use lightweight datasets JDB~\cite{sun2023journeydb} (generation) and ShareGPT4V\_PT~\cite{chen2024sharegpt4v} (understanding)
, together for a single epoch with accumulated batch size=128 and learning rate=5e-5.
We then apply a lightweight, single-epoch fine-tuning on ShareGPT4V for one epoch
with accumulated batch size=256 and learning rate=1e-4.
Unless otherwise specified, all results are obtained with this training schedule. We use the original tokenizer in the unified model.

\begin{figure*}[t]
  \centering
  \includegraphics[width=0.95\linewidth]{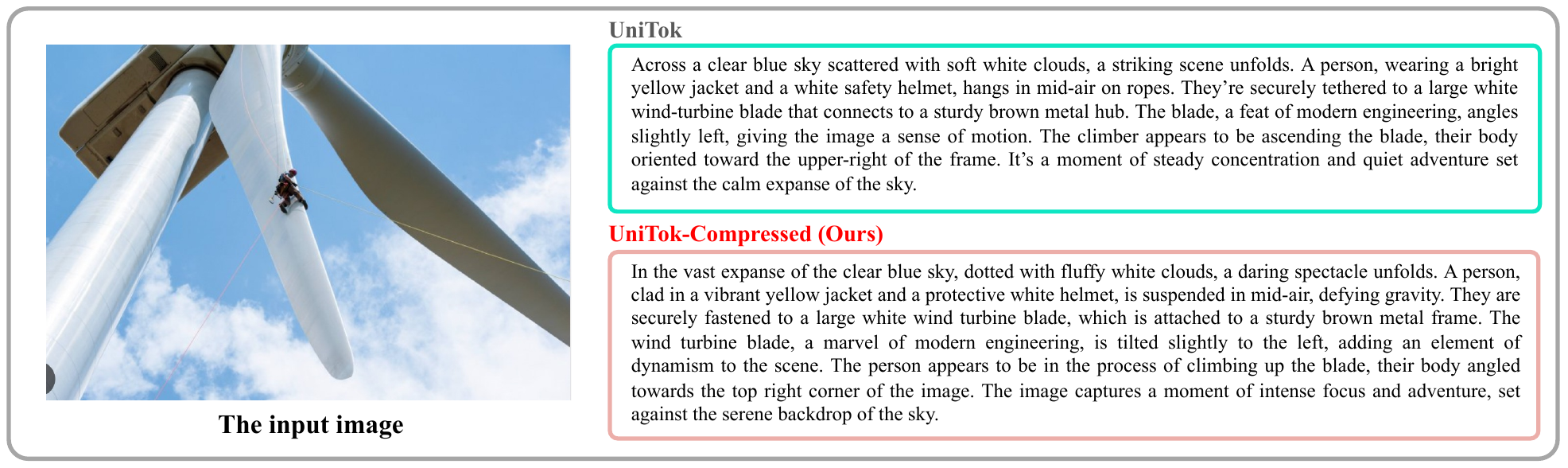}
  \vspace{-0.2cm}
  \caption{Understanding task examples: generating the texts that describe the image.}
  \label{fig:under_ex}
\end{figure*}

\paragraph{Baselines.}
We compare 6 representative, high-performing
unified model backbones 
and their compressed variants: \textsc{UniTok}~\cite{ma2025unitok}, \textsc{Vila-U}~\cite{wu2024vila}, \textsc{VARGPT}~\cite{zhuang2025vargpt}, \textsc{UniFork}~\cite{li2025unifork}, and \textsc{OpenUni}~\cite{wu2025openuni}, and \textsc{BAGEL}~\cite{deng2025emerging}.
Among these methods, \textsc{UniTok}, \textsc{Vila-U}, and \textsc{VARGPT} use one image tokenizer for both understanding and generation tasks, which is the main focus of our {\n}. Beyond them, we also plugin-and-play more diverse baselines using {\n}, including different tokenizers for understanding and generation (\ie \textsc{UniFork}), and unified model with diffusion model (\ie \textsc{OpenUni} and \textsc{BAGEL}).
For each unifined model, we create a \emph{compressed} version by inserting our plug-in stack (global extractor, compressor, decompressor) into their vision tokenizer (and diffusion model, if applicable) while keeping the language model interface unchanged.
Our setup here uniformly refers to the versions of each unified model that use 256 image tokens (except for \textsc{OpenUni} and Bagel, since they use diffusion models), and that use Llama-3.2-1B as the language model.
Unless stated otherwise, we set the downsampling factor to \(s=2\) (i.e., \(4\times\) fewer local tokens) and use \(N_g=4\) global tokens, which our ablation identifies as the best accuracy–efficiency trade-off.
There is no prior unified model compression baseline that jointly supports understanding and generation under a single autoregressive interface; therefore we report each unified model against its own \emph{compressed} counterpart.

\paragraph{Benchmarks.}
We evaluate on standard multimodal benchmarks—GQA~\cite{hudson2019gqa}, MME~\cite{fu2023mme}, POPE~\cite{li2023evaluating}, Seed-bench~\cite{li2024seed},TextVQA~\cite{antol2015vqa}, MMMU~\cite{yue2024mmmu}, and MM-Bench~\cite{liu2024mmbench} using the official splits and metrics
All methods share the same input budgets and decoding hyperparameters.
For understanding, the language model directly consumes the continuous visual tokens \(\{\mathbf{G}, \hat{\mathbf{X}}\}\) produced by the enhanced tokenizer, without quantization, delimited in the prompt by \texttt{[IMG\_BOS]} \dots \texttt{[IMG\_SEP]} \dots \texttt{[IMG\_EOS]}. We evaluate the generation performance on MJHQ-30K dataset~\cite{li2024playground}. We report Fréchet Inception Distance (FID)~\cite{heusel2017gans}, which compares the means and covariances of Inception-feature distributions between generated and real images; lower values indicate better quality. We also report CLIPScore~\cite{radford2021learning} (from CLIP), the cosine similarity between text and image embeddings in CLIP’s joint space, which measures alignment between images and text; higher values indicate better alignment.
For generation, the language model predicts compressed-domain indices for \emph{both} global and local tokens, which are then de-quantized and decompressed into dense features before the image decoder.
For methods without a diffusion model (i.e., \textsc{UniTok}, \textsc{Vila-U}, \textsc{VARGPT}, and \textsc{UniFork}), we use identical sampling procedures and inference budgets (same guidance, number of steps, and temperature) to ensure comparability.
For methods with a diffusion model (i.e., \textsc{OpenUni} and \textsc{BAGEL}), we keep the same downsampling factor \(s=2\), while applying slightly different compression schemes to the diffusion model’s input and output sizes.

\begin{figure*}[t]
  \centering
  \includegraphics[width=0.8\linewidth]{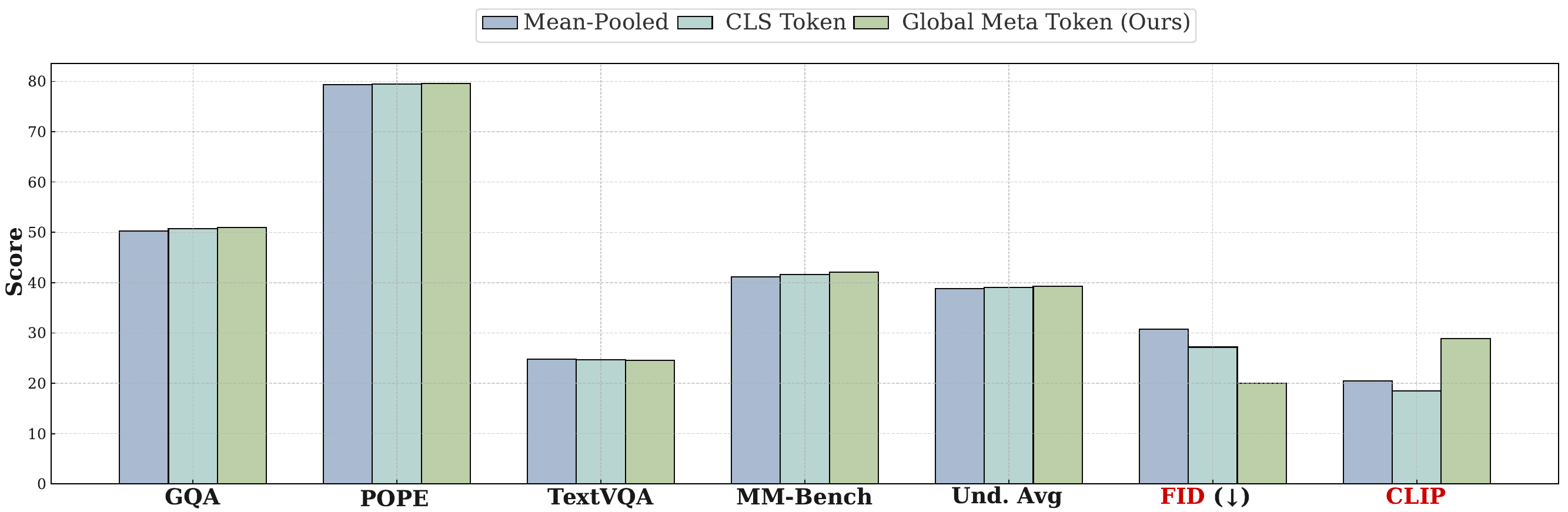}
  \vspace{-0.2cm}
  \caption{Ablation on global token type. Results use $N_g{=}4$. Our global meta token yields competitive understanding and notably stronger generation quality (lower FID, higher CLIP).}
  \label{fig:global_type}
\end{figure*}

\subsection{Vision-Language Understanding Performance}
Table~\ref{tab:understanding-performance} reports results on standard visual understanding benchmarks, comparing each baseline with its compressed counterpart. Across methods, our compression framework consistently maintains strong accuracy despite substantial reductions in visual token length. For example, \textsc{UniTok-Compressed} shows only minor drops on GQA (55.71$\rightarrow$53.07) and POPE (82.66$\rightarrow$79.36), while largely preserving performance on MME and MMMU. For baselines that incorporate diffusion models, {\n} likewise preserves understanding quality: Seed-bench within \textsc{OpenUni} decreases only slightly (48.39$\rightarrow$47.51), and \textsc{OpenUni-Compressed} even outperforms the original on MM-Bench. These findings confirm the robustness of our globally guided decompression design, which enables accurate visual--semantic reconstruction from compact representations. A further advantage of our approach is its modular, plug-in nature: without modifying backbones or introducing additional supervision, the same token compressor can be seamlessly applied across architectures, highlighting its generality for transformer-based vision-language systems.

Figure~\ref{fig:under_ex} compares captions produced from the same LLM when using dense \textsc{UniTok} tokens versus our pooled, globally guided compressed tokens. The compressed variant preserves the key entities and relations in the scene (person, wind turbine, blue sky) while maintaining spatial layout and action cues (climbing direction, body orientation). This example illustrates that understanding remains robust under pooling-based token compression, consistent with our quantitative trends.

\subsection{Image Generation Performance}
Table~\ref{tab:generation-performance} reports FID (lower is better) and CLIP similarity (higher is better) for each baseline and its compressed variant. Overall, compression tends to increase FID moderately while often reducing CLIP alignment, though the magnitude varies by backbone. 

On lighter backbones, the degradation is small: \textsc{UniTok} changes from 16.14/30.5 (FID/CLIP) to 16.33/22.0, and \textsc{VARGPT} from 14.77/24.2 to 15.02/21.6. \textsc{Vila-U} is comparatively robust, with FID 14.80$\rightarrow$16.37 and CLIP 29.8$\rightarrow$28.9. Among stronger models, \textsc{UniFork} shows a slight CLIP \emph{increase} (25.5$\rightarrow$26.0) despite FID rising to 20.24. \textsc{BAGEL} attains the best baseline quality (12.73 FID, 32.0 CLIP); its compressed version remains competitive at 17.22 FID and 28.8 CLIP. The largest drop is observed for \textsc{OpenUni} (16.45/26.7$\rightarrow$24.29/22.3), indicating greater sensitivity to token reduction for that design.
On lighter backbones, the degradation is small: \textsc{UniTok} changes from 16.14/30.5 (FID/CLIP) to 16.33/22.0, and \textsc{VARGPT} from 14.77/24.2 to 15.02/21.6. \textsc{Vila-U} is comparatively robust, with FID 14.80$\rightarrow$16.37 and CLIP 29.8$\rightarrow$28.9. Among stronger models, \textsc{UniFork} shows a slight CLIP \emph{increase} (25.5$\rightarrow$26.0) despite FID rising to 20.24. \textsc{BAGEL} attains the best baseline quality (12.73 FID, 32.0 CLIP); its compressed version remains competitive at 17.22 FID and 28.8 CLIP. The largest drop is observed for \textsc{OpenUni} (16.45/26.7$\rightarrow$24.29/22.3), indicating greater sensitivity to token reduction for that design.

Taken together, these results show that, while token compression introduces some loss in FID, many models retain strong semantic alignment (CLIP), and several backbones (e.g., \textsc{Vila-U}, \textsc{VARGPT}, \textsc{UniFork}) remain close to their full-token counterparts, supporting the viability of compressed visual inputs for high-quality image synthesis.

\begin{figure*}[t]
  \centering
  \includegraphics[width=0.9\linewidth]{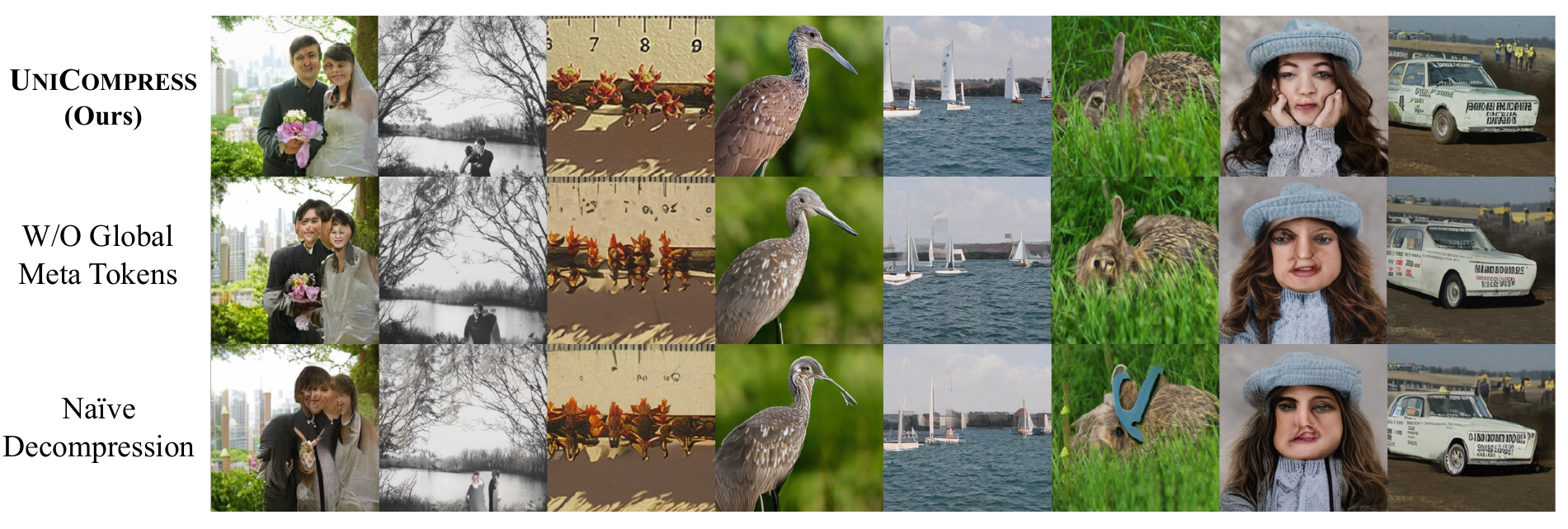}
  \vspace{-0.2cm}
  \caption{ {\n} preserves the most visual information under compression by using global meta tokens and autoregressive decompressor.}

  \label{fig:tokenizer_ex}
\end{figure*}

\subsection{Training and Inference Efficiency} Table~\ref{tab:efficiency} summarizes the training and inference efficiency across all models. Compressed variants consistently reduce training time and inference latency, with the largest gain observed in generation settings. In particular, \textsc{UniTok-Compressed} reduces generation inference time from 32.25 minutes to 18.96 minutes, corresponding to a relative speedup of over 40\%. Similar improvements are observed in \textsc{Vila-U} and \textsc{VARGPT}. These results highlight a major advantage of our approach: it enables end-to-end acceleration of generation pipelines, which has been difficult to achieve in prior token compression frameworks. Most existing methods primarily optimize training throughput or reduce memory usage, but show limited effect on actual decoding time. By contrast, our framework improves runtime efficiency without compromising task performance. The benefits of compression are especially pronounced in autoregressive generation, where each token directly impacts latency. These findings demonstrate the practical utility of our method for real-world deployment, particularly in scenarios with limited compute budgets. 

\begin{figure}[t]
  \centering
  \includegraphics[width=0.85\linewidth]{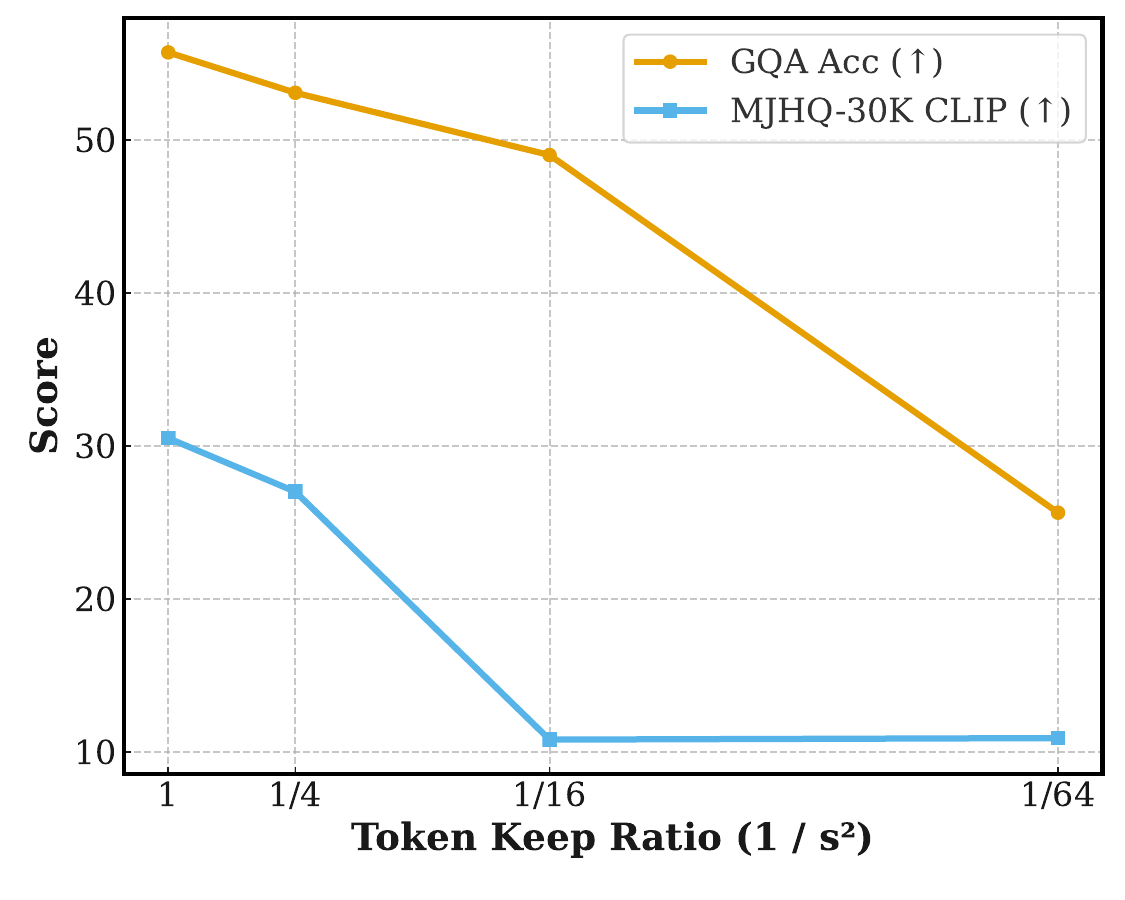}
  \vspace{-0.2cm}
  \caption{Effect of token keep ratio on accuracy. GQA (understanding) vs.\ MJHQ-30K CLIP (generation).}
  \label{fig:compression_ratio}
\end{figure}

\begin{table}[htp]
\centering
\caption{Ablation on local token compression (pooling/selection). All rows target the same token budget (\(\times4.0\)) as our default setting with \(s=2\).
Results use \(N_g=4\).}
\resizebox{0.48\textwidth}{!}{
\begin{tabular}{l|cccc|c}
\toprule
\textbf{Compressor} & \textbf{GQA} & \textbf{TextVQA} & \textbf{MM-Bench} & \textbf{Und. Avg} & \textbf{FID($\downarrow$)} \\
\midrule
AvgPool (Ours) & 53.07 & 24.66 & 42.14 & 39.29 & 20.01 \\
MaxPool & 50.80 & 24.40 & 41.90 & 39.03 & 20.15 \\
Strided Conv ($2{\times}2$) & 51.20 & 24.70 & 42.10 & 39.33 & 19.95 \\
Top-$k$ Token Selection & 49.90 & 24.00 & 41.20 & 38.37 & 21.00 \\
Learned Pooling (Gated) & 51.30 & 24.75 & 42.20 & 39.42 & 19.90 \\
\bottomrule
\end{tabular}
}
\label{tab:abl_pooling}
\end{table}

\subsection{Ablation Studies} 

\paragraph{Effect of Token Compression Ratio.}
We study how the token \emph{keep ratio} ($1/s^{2}\!\in\!\{1,\,1/4,\,1/16,\,1/64\}$ with stride $s\!\in\!\{1,2,4,8\}$) affects performance by compressing the $H{\times}W$ token grid via non-overlapping average pooling. This design preserves spatial layout, hence only integer pooling windows (e.g., $2{\times}2$, $4{\times}4$) are used. As shown in Fig.~\ref{fig:compression_ratio}, \textit{(1)} we use pooling-based compression to retain spatial structure, therefore the ratios are restricted to $1/s^{2}$ with $s\in\{2,4,\dots\}$; \textit{(2)} understanding is relatively robust to compression (GQA drops moderately from 55.71 at $1$ to 49.00 at $1/16$), while generation is far more sensitive (MJHQ\mbox{-}30K CLIP falls sharply from 30.5 to $\sim$11), validating our introduction claim; \textit{(3)} a keep ratio of $1/4$ ($s{=}2$) provides a good trade-off. Further compression yields noticeable degradation, especially on generation metrics.

\paragraph{Global token type.}
We compare three ways to form global tokens: mean-pooled image tokens, a CLS token from a ViT encoder, and our learnable global meta tokens.
As shown in Fig.~\ref{fig:global_type}, understanding metrics (e.g., GQA, TextVQA, MM-Bench, and the averaged score) stay close across all choices, indicating that coarse global summarization is generally sufficient for recognition-style tasks. In contrast, generation quality differs markedly: 
our global meta tokens achieve substantially lower FID and higher CLIP than mean-pooling or a CLS token. We attribute this gap to the query-based extraction that explicitly “reads” the whole token map and writes image-specific global semantics, which provides stronger conditioning for the decompressor. Hence, while all types are adequate for understanding, the learnable global meta tokens are crucial to preserve fidelity in generation. We also evaluate the impact of global token number in 

Figure~\ref{fig:tokenizer_ex} visualizes the tokenizer compression stage before any LLM decoding.
From top to bottom: ours ({\n} with autoregressive decompressor), ablation without global meta tokens, and naïve decompression (non-autoregressive). Under the same keep ratio, {\n} retains the most visual detail and global structure; removing global tokens breaks long-range consistency, while dropping the autoregressive decompressor yields over-smoothed textures and artifacts. This confirms that global guidance and autoregressive decompression are both necessary to preserve fidelity under compression.

\section{Conclusion}

We present {\n}, a token compression framework for unified models that supports both understanding and generation. The method adds a lightweight, modular compression–decompression mechanism centered on global token guided reconstruction, which reduces the number of visual tokens while preserving task performance. With a two stage training pipeline that keeps the language model unchanged, {\n} can be integrated into existing systems without full model retraining. Experiments across diverse benchmarks show up to $4\times$ reduction in visual tokens and over $40\%$ faster generation inference, with minimal loss in accuracy or image quality. In contrast to prior approaches that emphasize only training efficiency or understanding tasks, {\n} delivers consistent gains in both training and inference and for understanding and generation. These results indicate that compact visual representations can enable practical deployment under limited compute and memory, and they point toward scalable unified models that maintain quality while operating at much lower budgets.

{
    \small
    \bibliographystyle{ieeenat_fullname}
    \bibliography{main}
}
\appendix
\newpage

\begin{figure*}[t]
    \centering
    \includegraphics[width=\textwidth]{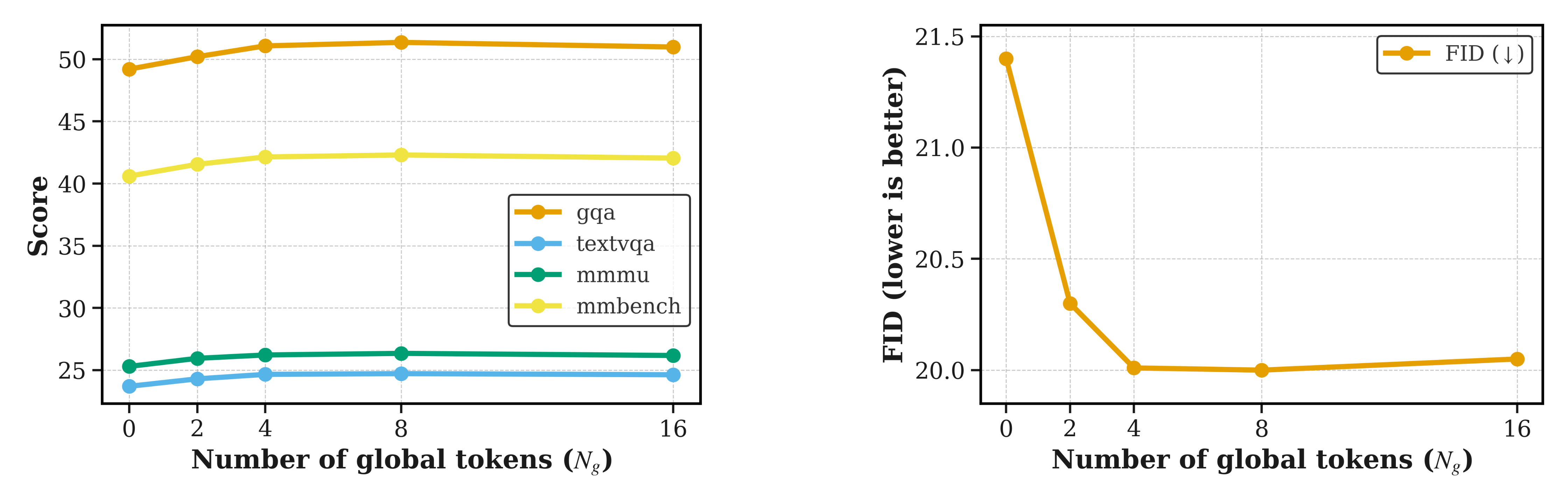}
    \caption{Effect of the number of global tokens $N_g$. Left: vision–language understanding (GQA, TextVQA, MMMU, MMBench; higher is better). Right: image generation quality (FID; lower is better).}
    \label{fig:ng_ablation}
\end{figure*}

\section{Additional Experimental Results}

\paragraph{System.}
We conduct our experiments on a single-node Ubuntu 22.04 LTS server equipped with an Intel(R) Xeon(R) Platinum 8468V CPU and 8$\times$ NVIDIA H100 80GB GPUs. Unless otherwise noted, all training and inference run on this machine using multi-GPU data parallelism.

\paragraph{Number of global tokens.}
As shown in Figure \ref{fig:ng_ablation}, we ablate \(N_g\in\{0,2,4,8,16\}\) and find a clear sweet spot at \(N_g=4\).
Removing globals (\(N_g=0\)) consistently hurts both understanding and generation (e.g., higher FID \(\approx 21.4\)).
Introducing a small set (\(N_g=2\)) improves all metrics but still trails larger settings.
Performance largely plateaus for \(N_g\in\{4,8\}\): understanding scores at 8 are only marginally above 4 (often within \(\sim 0.2\!-\!0.3\) absolute), and FID at 4 (\(\approx 20.01\)) matches 8 (\(\approx 20.00\)).
Since \(N_g\) increases sequence length and compute roughly linearly, we adopt \(N_g=4\) as the best accuracy–efficiency trade-off: it recovers global semantics sufficiently to guide decompression, reaches near-peak accuracy, and avoids the diminishing returns observed at \(N_g\ge 8\).


\end{document}